\documentclass{article}

\usepackage{arxiv}

\usepackage[utf8]{inputenc} 
\usepackage[T1]{fontenc}    
\usepackage{hyperref}       
\usepackage{url}            
\usepackage{booktabs}       
\usepackage{amsfonts}       
\usepackage{nicefrac}       
\usepackage{microtype}      
\usepackage{lipsum}
\usepackage{graphicx}
\graphicspath{ {./images/} }

\usepackage{color}
\usepackage{soul}
\usepackage{algpseudocode}
\usepackage{comment}
\usepackage{wrapfig}
\usepackage{algorithm}
\usepackage{amsmath,amssymb,amsthm}

\usepackage{multirow}
\usepackage{makecell}
\usepackage{mathrsfs}
\usepackage{varioref}
\usepackage{cleveref}
\usepackage{xspace}
\usepackage{bm}
\usepackage{enumitem}
\usepackage{pifont}
\usepackage{mathtools}
\usepackage{natbib}

\usepackage{subcaption}

\title{\textsc{MeTA-LoRA}: Data-Efficient Multi-Task Fine-Tuning for Large Language Models}

\author{
 Bo Cheng\thanks{Equal contribution} \\
  School of Artificial Intelligence, Jilin University\\
  \texttt{chengbo9691@gmail.com} \\
   \And
 Xu Wang\footnotemark[1] \\
  School of Artificial Intelligence, Jilin University\\
\texttt{xwang22@mails.jlu.edu.cn}\\
   \And
 Jinda Liu \\
  School of Artificial Intelligence, Jilin University\\
\texttt{liujd9922@mails.jlu.edu.cn}\\
\And
 Yi Chang \\
  School of Artificial Intelligence, Jilin University\\
  Engineering Research Center of Knowledge-Driven Human-Machine Intelligence, MOE, China\\
  International Center of Future Science, Jilin University\\
\texttt{yichang@jlu.edu.cn}\\
  \And
 Yuan Wu\textsuperscript{\thanks{ Corresponding author}} \\
  School of Artificial Intelligence, Jilin University\\
  \texttt{yuanwu@jlu.edu.cn} \\
}

\begin{document}
\maketitle

\begin{abstract}
Low-Rank Adaptation (LoRA) has emerged as one of the most widely used parameter-efficient fine-tuning (PEFT) methods for adapting large language models (LLMs) to downstream tasks. While highly effective in single-task settings, it struggles to efficiently leverage inter-task knowledge in complex multi-task learning scenarios, often requiring substantial task-specific data to achieve optimal performance. To address this limitation, we introduce \textsc{MeTA-LoRA}, a two-stage optimization framework that significantly improves data efficiency in multi-task adaptation. In the first stage, task-specific LoRA adapters are learned using only a few samples from each involved dataset, enabling rapid adaptation without large-scale supervision. In the second stage, the shared LoRA adapter is updated by aggregating gradients from multiple tasks to promote knowledge transfer across tasks, further reducing data usage by leveraging common patterns. In both multi-task learning and multilingual learning scenarios, our method matches or surpasses the performance of traditional full-data LoRA fine-tuning approaches, while using significantly less task-specific data.
\end{abstract}

\section{Introduction}
Large language models (LLMs) have transformed natural language processing by achieving state-of-the-art results on tasks from text generation to complex reasoning \citep{brown2020language,devlin2019bert,chang2024survey}. 
However, the sheer scale of these models, which often encompass billions of parameters, renders full-parameter fine-tuning prohibitively expensive in both computational and memory requirements, especially when adapting to multiple tasks simultaneously~\citep{han2024parameter}. As real-world applications increasingly demand multi-task capabilities, methods that reduce resource overhead while preserving performance have become critical. Parameter-efficient fine-tuning (PEFT) techniques~\citep{hu2022lora,rebuffi2017adapters,li2021prefix,chang2024ba}, which add only lightweight adaptation modules to a frozen base model, offer a promising solution by slashing trainable parameters from hundreds of millions to mere thousands dramatically cutting GPU memory footprint and speeding up training without sacrificing modularity across tasks.

While PEFT approaches like LoRA~\citep{hu2022lora} and its multi-task extensions (e.g., R-LoRA~\citep{liu2025rlora}, HydraLoRA~\citep{tian2024hydralora}) deliver substantial efficiency gains in single-task scenarios, they still rely on large volumes of labeled data when scaled to many tasks resulting in poor data efficiency. For example, HydraLoRA reports that fine-tuning on just 43 tasks required over 320,000 in-domain examples to achieve satisfactory performance on downstream tasks~\citep{tian2024hydralora}. This heavy data demand undermines the very efficiency gains PEFT seeks to provide in multi-task settings and highlights the pressing need for methods that balance both parameter- and data-efficiency in large-scale multi-task adaptation. 

Data‐efficient techniques like coreset selection~\citep{liu2023coreset1,xia2024less} or data pruning~\citep{azeemi2023dataprunig} excel at trimming down data for a single task by homing in on the “most informative” examples, but in a multi‐task LLM scenario, this narrow focus comes at the expense of broader generalization across many objectives. By optimizing for task-specific highlights, these methods tend to under-represent the shared structures and cross-task patterns that are essential for robust performance on unseen or less frequent tasks~\citep{lin2024re,chen2024molecular}.

Inspired by the principles of Meta-Learning, which emphasizes enabling models to “learn how to learn”~\citep{finn2017maml}, we propose a novel framework called \textsc{MeTA-LoRA}, specifically designed to address the data efficiency challenge in the fine-tuning process of LLMs within multi-task learning scenarios. To achieve this, we frame the fine-tuning process as an iterative optimization procedure consisting of two key stages: task-specific adaptation and meta-knowledge update. The task-specific adaptation stage allows for rapid per-task learning using only a small amount of data from the support set, enabling efficient adaptation to each task without requiring large datasets. Meanwhile, the meta-knowledge update stage aggregates insights from multiple tasks through a shared LoRA adapter, promoting the transfer of knowledge across tasks while minimizing data usage. Together, these two stages optimize the learning process, significantly enhancing data efficiency while preserving model performance. We demonstrate that \textsc{MeTA-LoRA} achieves competitive performance in both multi-task learning and multilingual learning scenarios while requiring significantly less task-specific data, showcasing its adaptability across a variety of tasks. Extensive ablation studies and analyses further demonstrate the necessity of the two-stage optimization framework.

In summary, the major contributions of this paper are outlined below.

\begin{itemize}
    \item We propose \textsc{MeTA-LoRA}, a novel framework designed to enhance data efficiency for multi-task LoRA adaptation while preserving model performance.
    \item Comprehensive experimental evaluations validate the effectiveness of \textsc{MeTA-LoRA} in both multi-task learning and multilingual learning scenarios.
    \item Extensive ablation experiments and analyses demonstrate the necessity of the two-stage optimization framework.
\end{itemize}

\section{Related Work}

\paragraph{Parameter-Efficient Fine-tuning} As LLMs grow more powerful, fine-tuning them remains computationally intensive. This challenge has motivated the development of parameter-efficient fine-tuning (PEFT) methods, which aim to lower memory and storage demands during model adaptation. One representative approach is adapter tuning~\citep{rebuffi2017learning, houlsby2019parameter,sung2022vl, stickland2019bert}, which introduces trainable layers into the existing model while keeping the original parameters frozen. Another line of PEFT research focuses on directly manipulating model activations through learnable vectors, with methods such as concatenation~\citep{liu2024gpt, li2021prefix, lester2021power}, multiplication, and addition.
Additionally, prompt-based tuning methods like prefix tuning~\citep{lester2021power} and continuous prompt tuning~\citep{li2021prefix, liu2021p} replace discrete prompt engineering with trainable embeddings.
Beyond injecting new parameters, researchers have also explored sparse updates~\citep{sung2021training, dey2024sparse} and low-rank adaptation (LoRA)~\citep{hu2022lora} as alternatives that modify only a small subset of the model’s parameters or its computational graph.

\paragraph{Lora Architecture on Multi-Task Learning} Multi-LoRA Architectures have emerged as a promising solution for adapting LLMs, such as LLaMA, in resource-constrained settings~\citep{touvron2023llama}. To further leverage the potential of LoRA, researchers have proposed multi-LoRA approaches that employ multiple low-rank adapters simultaneously. For instance, LoRAHub~\citep{huang2023lorahub}  trains multiple adapters and dynamically selects suitable combinations based on the domain at inference time, while MultiLoRA~\citep{wang2023multilora} improves scalability by decomposing LoRA modules and introducing learnable scaling factors. To reduce resource usage, LoRAFit~\citep{zadouri2023pushing} refines module structure, and LoRAMoE~\citep{dou2023loramoe} incorporates a Mixture-of-Experts framework to protect pre-trained knowledge during instruction tuning.

\paragraph{Data-efficient strategies on LLM} Fine-tuning LLMs is computationally demanding, which has motivated the development of data-efficient strategies to reduce resource consumption without sacrificing performance~\citep{ding2023efficiency, xu2024survey}. Among these, two prominent approaches are coreset selection~\citep{lin2024data, xia2022moderate, xia2024less} and data pruning~\citep{marion2023less, sorscher2022beyond}. These techniques highlight the promise of intelligent data selection for alleviating the computational burden of LLM fine-tuning, thereby improving accessibility in resource-constrained settings. Specifically, coreset selection focuses on identifying a representative subset of training data that can approximate the performance achieved with the full dataset~\citep{lin2024data, xia2022moderate, xia2024less}, while data pruning aims to eliminate less informative or redundant samples to streamline training~\citep{marion2023less, sorscher2022beyond}.
\section{\textsc{MeTA-LoRA}}
\label{sec:meta_lora}

\begin{wrapfigure}[31]{r}{0.47\textwidth} 
    \vspace{-5pt} 
    \centering
    \includegraphics[width=\linewidth]{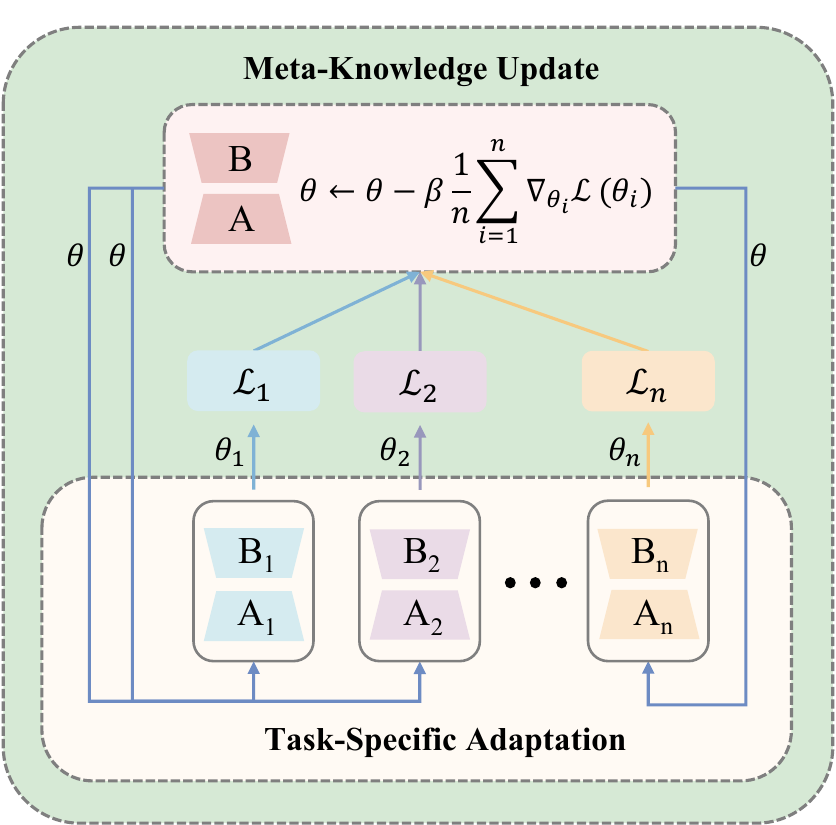}
    \caption{\textbf{Architecture of \textsc{MeTA-LoRA}}. During fine-tuning, \textsc{MeTA-LoRA} adopts a two-stage optimization framework comprising a task-specific adaptation stage and a meta-knowledge update stage. Starting from the initial shared LoRA parameters $\theta$, task-specific adapters are rapidly adapted using support set of each task, enabling efficient task-level specialization. In the second stage, the updated task-specific parameters ${\theta_1, \theta_2, \dots, \theta_n}$ are used to compute gradients from the corresponding query sets, which are then aggregated to update the shared LoRA adapter, enabling effective cross-task knowledge transfer. By iteratively alternating between these two stages, \textsc{MeTA-LoRA} enhances data efficiency in multi-task adaptation of LLMs.}
    \label{method_pic}
\end{wrapfigure}

In this section, we introduce \textsc{MeTA-LoRA}, a data-efficient LoRA-based architecture for fine-tuning, as illustrated in Figure~\ref{method_pic}. The detailed fine-tuning procedure is provided in Algorithm~\ref{alg:metalora}.

\subsection{Problem Formulation}
Given a model $f$ with pre-trained weights $W_0$ and a set of tasks \( \mathcal{T} = \{ \mathcal{T}_1, \mathcal{T}_2, \dots, \mathcal{T}_N \} \), current LoRA-based methods for multi-task adaptation optimize:
\begin{equation}
    \min_{\theta} \sum_{i=1}^N \mathbb{E}_{(x,y)\sim \mathcal{T}_i} [\mathcal{L}(f_{W_0+\Delta{\theta}}(x), y)]
\end{equation}
where $\theta=\{A, B\}$ denotes the shared LoRA adapter parameters across all $N$ tasks, and $\Delta{\theta}=BA$ approximates the accumulated gradient updates $\Delta{W_0}$. $\mathcal{T}_i$ denotes the data distribution of the $i$-th task. Different from the above paradigm, we propose a two-stage optimization framework that maintains multiple local LoRA adapters during the task-specific adaptation stage, alongside a global LoRA adapter updated in the meta-knowledge update stage, enabling the model to efficiently adapt to multiple tasks while minimizing the amount of data required for each task. 


\subsection{Fine-tuning}

\subsubsection{Phase I: Task-Specific Adaptation}

In each iteration, we randomly sample $n$ tasks $\mathcal{B} = \{\mathcal{T}_1, \mathcal{T}_2, \dots, \mathcal{T}_n\}$ from $\mathcal{T}$ with each task following the episodic formulation:
\begin{equation}
    \mathcal{T}_i \coloneqq (\mathcal{S}_i, \mathcal{Q}_i)
\end{equation}
where $\mathcal{S}_i$ and $\mathcal{Q}_i$ are disjoint subsets randomly selected from task $\mathcal{T}_i$, traditionally referred to as the support set and the query set~\citep{finn2017maml} with sizes $n_i^s$ and $n_i^q$, respectively.

For each task $\mathcal{T}_i \in \mathcal{B}$, the task‐specific adapter parameters $\theta_i$ are first initialized with the shared adapter parameters $\theta$ (i.e., $\theta_i \gets \theta$). Subsequently, the model performs the gradient descent on $\theta_i$ for $k$ steps based on its corresponding support set, simulating the model's rapid adaptation to a new task. One gradient update of task-specific (local) LoRA adapter parameters $\theta_i$ can be formulated as:
\begin{equation}
\theta_i \gets \theta_i - \alpha \nabla_{\theta_i} \mathcal{L}_{\mathcal{S}_i}({\theta_i})
\label{eq:inner_update}
\end{equation}
where $\alpha$ is the learning rate of the adaptation stage, and the task-specific adaptation loss $\mathcal{L}_{\mathcal{S}_i}(\theta_i)$ is:

\begin{equation}
\mathcal{L}_{\mathcal{S}_i}(\theta_i) = \frac{1}{n_i^s} \sum_{(x,y)\sim \mathcal{S}_i}\mathcal{L}(f_{W_0 + \Delta \theta_i}(x), y)
\label{eq:inner_loss}
\end{equation}

\begin{algorithm}[!t]
\caption{\textsc{MeTA-LoRA} using the First-Order Approximation}
\label{alg:metalora}
\begin{algorithmic}[1]
\State \textbf{Require:} Task set \( \mathcal{T} = \{\mathcal{T}_1, \mathcal{T}_2, \dots, \mathcal{T}_N\} \), model \( f \) with frozen pre-trained parameters \( W_0 \),  learning rate of the adaptation stage $\alpha$, learning rate of the meta-update stage $\beta$, number of adaptation steps $k$, number of selected tasks in each iteration $n$
\State Randomly initialize the shared LoRA adapter parameters $\theta$
\For{each iteration}
    \State \textbf{Stage I: Task-Specific Adaptation}
    \State Randomly select $n$ tasks \( \mathcal{B} \subseteq \mathcal{T} \)
    \For{each task \( \mathcal{T}_i \in \mathcal{B} \)}
        \State Sample support set \( \mathcal{S}_i \) and query set \( \mathcal{Q}_i \) 
        \State Create a temporary copy of parameters for adaptation: $\theta_i \leftarrow \theta$
        \State Obtain task-specific adapted parameters \( \theta_i \) by performing Eq.~\ref{eq:inner_update} for $k$ steps
    \EndFor
    \State \textbf{Stage II: Meta-Knowledge Update}
    \State Compute the generalization loss for each selected task \( \mathcal{T}_i \) using Eq.~\ref{ea:outer_loss}
    \State Average the generalization gradients to update the shared LoRA parameters \( \theta \) using Eq.~\ref{eq:outer_update}
\EndFor
\State \textbf{Output:} Fine-tuned LoRA parameters \( \theta \)
\end{algorithmic}
\end{algorithm}

\subsubsection{Phase II: Meta-Knowledge Update}
After obtaining the adapted parameters $\theta_i$ for each task $\mathcal{T}_i$, the meta-update is performed on the shared (global) LoRA parameters $\theta$. To maintain computationally tractable with large models, we employ the first-order approximation of MAML, a common simplification that retains meta-learning signals while avoiding the prohibitive computational overhead of Hessian-vector products. In practice, this is implemented by detaching the computational graph of the task-specific adaptation, which prevents backpropagation through the task-specific update steps. The approximate meta-update rule, which averages the gradients from a batch of $n$ sampled tasks, is therefore formulated as:
\begin{equation}
\theta \gets \theta - \beta \frac{1}{n} \sum_{i=1}^n  \nabla_{\theta_i} \mathcal{L}_{\mathcal{Q}_i}(\theta_i)
\label{eq:outer_update}
\end{equation}
where $\beta$ is the learning rate in the meta-stage, and the generalization loss $\mathcal{L}_{\mathcal{Q}_i}(\theta_i)$ with respect to task $\mathcal{T}_i$ is computed on the corresponding query set using the adapted LoRA parameters $\theta_i$:
\begin{equation}
    \mathcal{L}_{\mathcal{Q}_i}(\theta_i) = \frac{1}{n_i^q} \sum_{(x,y)\sim \mathcal{Q}_i} \mathcal{L}(f_{W_0 + \Delta \theta_i}(x), y)
\label{ea:outer_loss}
\end{equation}

\subsection{Inference}
After fine-tuning, we obtain a single shared adapter that captures multi-task knowledge. It is worth emphasizing that the task-specific adapters are merely intermediate artifacts generated during the fine-tuning process. In the inference stage, the parameters of the shared adapter $\theta$ are seamlessly merged into the frozen pre-trained weights $W_0$, eliminating the need to activate or switch among task-specific adapters. 
\section{Experiments}
In this section, we present a series of experiments to evaluate the effectiveness of \textsc{MeTA-LoRA} in both multi-task and multilingual learning scenarios. Additionally, we conduct ablation studies to examine the necessity of the proposed two-stage optimization framework, as well as the adaptability of \textsc{MeTA-LoRA} under more challenging multi-task configuration. Finally, we perform a sensitivity analysis to investigate the impact of the amount of fine-tuning data on overall performance.

\subsection{Experiment Setting}
\subsubsection{Multi-task Learning}
In multi-task learning scenario, we consider two task configurations. First, we adopt the same experimental setting described in HydraLoRA~\citep{tian2024hydralora}. Additionally, we develop a five-task dataset comprising semantically and structurally diverse samples, to further evaluate the model's capacity to handle varied tasks. 

\textit{Flanv2 Setting}  

\textbf{Dataset.} 
We fine-tune the model using a subset of the Flanv2 dataset~\citep{wei2021flan-v2} that includes 43 tasks from both Natural Language Understanding (NLU) and Natural Language Generation (NLG), grouped into 10 distinct task clusters. The model’s performance is evaluated using the Big-Bench Hard (BBH)~\citep{suzgun2022bbh} benchmark. More details of the dataset can be found in Appendix~\ref{appendix:flanv2}. 

\textbf{Baselines.} 
We evaluate MeTA-LoRA against other LoRA-based methods designed for multi-task learning across multiple datasets: (1) Lorahub~\citep{huang2023lorahub}, which utilizes black-box optimization to learn weights for 20 randomly selected LoRAs for new tasks, applying weighted averaging without the need for gradient calculations; (2) LoRA MoE~\citep{liu2024moe}, which combines lightweight experts (LoRA) with a Mixture of Experts (MoE) architecture for high efficiency, enabling generalization to new tasks without prior knowledge; (3) HydraLoRA~\citep{tian2024hydralora}, which employs Multi-Head structure in conjunction with MoE to achieve a balance between parameter efficiency and training effectiveness.

\textit{Five-task Setting}

\textbf{Dataset.} 
We build a five-task dataset encompassing GSM8K~\citep{cobbe2021gsm8k} for arithmetic reasoning, QQP~\citep{wang2017qqp} and CosmosQA~\citep{huang2019cosmosqa} for NLU, SiQA~\citep{sap2019socialiqa} and PiQA~\citep{bisk2020piqa} for commonsense reasoning. Then, we use the MMLU (Massive Multitask Language Understanding)~\citep{hendrycks2020mmlu} benchmark to measure the world knowledge acquired during fine-tuning and problem solving ability of models. In addition, we include the BBH benchmark. Following common practice, we conduct 5-shot evaluation on MMLU and 3-shot evaluation on BBH.

\textbf{Baselines.} 
Following prior studies~\citep{tian2024hydralora,2025-malora}, we adopt LLaMA2-7B and LLaMA2-13B as the backbone models. For comparison, we evaluate \textsc{MeTA-LoRA} against two representative baselines: (1) LoRA, a widely used approach for parameter-efficient fine-tuning; (2) HydraLoRA, a state-of-the-art method tailored for multi-task learning.

\begin{table}[!t]
\caption{Performance of different methods on BBH (3-shot) with LLaMA2-7B and LLaMA2-13B fine-tuned on the subset of the Flanv2 dataset. \textsc{MeTA-LoRA} uses only 100 examples sampled from each task for fine-tuning, whereas the other methods except Base utilize the full dataset. * indicates results from \cite{tian2024hydralora}.}
\begin{center}
\arrayrulewidth=1pt
\renewcommand{\arraystretch}{1.2}
\resizebox{\textwidth}{!}{\begin{tabular}{m{55pt}<{\centering}|m{30pt}<{\centering}m{30pt}<{\centering}m{50pt}<{\centering}m{60pt}<{\centering}m{60pt}<{\centering}m{70pt}<{\centering}}
\hline
\textbf{Model Size} & \textbf{Base} & \textbf{LoRA} & \textbf{LoRAHub*} & \textbf{LoRA MoE*} & \textbf{HydraLoRA*} & \textbf{\textsc{MeTA-LoRA}} \\ 
\hline
7B & 31.6 & 37.2 & 39.7 & 40.3 & \textbf{41.8} & 38.52 \\ 
13B & 38.4 & 40.9 & 41.9 & 43.7 & 44.7 & \textbf{46.32} \\ 
\hline
\end{tabular}}
\end{center}
\label{main: Flanv2-results}
\end{table}

\begin{table*}[!t]
  \caption{Performance comparison on the proposed five-task dataset. The base models are fine-tuned by \textsc{MeTA-LoRA} with only 50 examples randomly sampled from each task, and LoRA and HydraLoRA represent the full-data tuning schemes.}
  \begin{center}
  \renewcommand{\arraystretch}{0.9}
  \small
  \resizebox{\textwidth}{!}{\begin{tabular}{m{70pt}<{\centering}|m{65pt}<{\centering}|m{65pt}<{\centering}|m{65pt}<{\centering}|m{65pt}<{\centering}}
    \toprule
    \textbf{Method} & \textbf{MMLU} & \textbf{MMLU-math} & \textbf{BBH}  & \textbf{AVG} \\
    \toprule
    
    \multicolumn{5}{c}{LLaMA2-7B} \\
    \midrule

    LoRA               & 29.43 & 26.42 & 30.24 & 28.70 \\
    HydraLoRA          & 46.61 & 30.00 & 37.49 & 38.03 \\
    \textsc{MeTA-LoRA} & 46.34 & 31.12 & 39.53 & \textbf{39.00} \\
    \midrule
    
    \multicolumn{5}{c}{LLaMA2-13B} \\
    \midrule

    LoRA               & 39.99 & 26.08 & 44.84 & 36.97 \\
    HydraLoRA          & 54.24 & 30.88 & 44.67 & 43.26 \\
    \textsc{MeTA-LoRA} & 54.18 & 30.97 & 46.12 & \textbf{43.76} \\
    
    \bottomrule
  \end{tabular}}
  \end{center}
  \label{main: 5-tasks-results}
\end{table*}

\subsubsection{Multilingual Learning}
In multilingual learning scenario, we fine-tune the LLaMA2-7B and LLaMA2-13B using Bactrian-X~\citep{li2023bactrian}, a comprehensive multilingual parallel dataset comprising 3.4 million instruction–response pairs across 52 languages. Bactrian-X is automatically constructed by translating instructions from Alpaca~\citep{taori2023alpaca} and Dolly~\citep{conover2023dolly} via the Google Translate API.2. In our evaluation, we compare \textsc{MeTA-LoRA} against two baselines: (1) the corresponding vanilla models; (2) the multilingual Bactrian models (BX), which are fine-tuned on the full Bactrian-X dataset. To probe the zero-shot language understanding capability of the different models and how much knowledge of they encode, we evaluate on the following four benchmarks:
\begin{itemize}
\item XCOPA~\citep{ponti2020xcopa}: a multilingual resource designed for causal commonsense reasoning, encompassing 11 languages from 11 families and several areas around the globe. The task requires selecting the correct subsequent sentence from two given options, based on cause and effect question types.
\item XStoryCloze~\citep{lin2022XStoryCloze}: the professionally translated version of the English StoryCloze dataset~\citep{mostafazadeh2016corpus} into 10 non-English languages. The task involves selecting one sentence as a plausible ending (closure) from two options, given a four-sentence story as the premise. 
\item XWinoGrad~\citep{XWinoGrad,XWinoGrad_crosslingual}: a multilingual assessment dataset for commonsense reasoning, made up of Winograd Schema Challenge problems in six languages. The objective is to select the most plausible sentence from options that differ slightly. 
\item EXAMS~\citep{hardalov2020exams}: A multilingual multiple-choice question-answering dataset constructed from high school exam questions in 16 languages, covering a wide range of subjects including natural sciences (e.g., physics), social sciences (e.g., history), and humanities (e.g., philosophy).
\end{itemize}

\subsubsection{Hyper-parameter Settings}

\begin{wraptable}{r}{0.6\linewidth} 
\vspace{-10pt} 
\centering
\caption{Statistics of different settings with respect to the number of tasks and the amount of samples used by the baselines and the proposed \textsc{MeTA-LoRA}.}
\label{tab:setting_sta}
\begin{tabular}{lccc}
\toprule
\textbf{Settings} & \textbf{Tasks} & \textbf{ Baselines} & \textbf{Ours} \\
\midrule
Flanv2    & 43 & 325,783 & 4,300 \\
Five-task  & 5 & 439,950 & 250 \\
Multilingual  & 52 & 3,400,000 & 2,600 \\
Five-task variant  & 5 & 487,075 & 250 \\
\bottomrule
\end{tabular}
\vspace{-8pt} 
\end{wraptable}

For both multi-task learning and multilingual learning scenarios, we run \textsc{MeTA-LoRA} for 1,000 iterations. In each iteration, we randomly select $2$ tasks (i.e. $n=2$) for adaptation. For each task $i$, the size of the support set $n_i^s$ and the query set $n_i^q$ are both set to $8$. We perform $3$ steps of gradient descent on the support set to obtain task-specific LoRA adapters, using a learning rate of $5 \times 10^{-6}$. During the meta-knowledge update phase, we employ AdamW optimizer with a learning rate of $2 \times 10^{-6}$. A comprehensive comparison of the number of fine-tuning samples used by the baselines and our method under these settings is presented in Table~\ref{tab:setting_sta}, and additional details regarding the hyper-parameter settings can be found in Appendix~\ref{appendix:paras}.


\subsection{Performance}
\subsubsection{Performance on Multi-task Learning}
Table~\ref{main: Flanv2-results} and Table~\ref{main: 5-tasks-results} present the performance of various fine-tuned models evaluated on the different benchmarks, using a subset of the Flanv2 dataset and the proposed five-task dataset, respectively. We make several observations in more detail, and discuss them below.
\begin{itemize}
    \item As shown in Table~\ref{main: Flanv2-results}, \textsc{MeTA-LoRA} consistently improves performance over both the base models and the models fine-tuned with standard LoRA. Notably, based on the LLaMA2-13B model, \textsc{MeTA-LoRA} achieves the highest BBH score of 46.32 among all LoRA-based baselines, despite involving only 100 examples per task into the fine-tuning process. The improvements highlight the effectiveness and data efficiency of the \textsc{MeTA-LoRA} framework. 
    \item When a more diverse set of tasks is used for fine-tuning, as shown in Table~\ref{main: 5-tasks-results}, \textsc{MeTA-LoRA} consistently outperforms both standard LoRA and HydraLoRA on LLaMA2-7B and LLaMA2-13B, even with only 50 examples per task. Overall, \textsc{MeTA-LoRA} surpasses HydraLoRA and LoRA by 0.97\% and 10.3\% on LLaMA2-7B, and by 0.50\% and 6.79\% on LLaMA2-13B, respectively, indicating that it can effectively capture task-specific knowledge and thereby enhances generalization across a broad range of tasks.
\end{itemize}

\begin{table}[!t]
\caption{Averaged zero-shot accuracy for XCOPA, XStoryCloze, XWinograd, and EXAMS under different tuning schemes. \textsc{MeTA-LoRA} fine-tunes the base models with only 50 examples randomly sampled from each language, and BX\textsubscript{LLaMA} represents LoRA tuning on the full Bactrian-X dataset. * indicates results from \cite{li2023bactrian}.}
\begin{center}
\small
\begin{tabular}{lccccc}
\toprule
\textbf{Model} & \textbf{XCOPA} & \textbf{XStoryCloze} & \textbf{XWinograd} & \textbf{EXAMS} & \textbf{AVG} \\
\midrule
LLaMA* (7B) & 50.22 & 57.03 & 57.96 & 28.20 & 48.35 \\
BX\textsubscript{LLaMA}* (7B) & 51.76 & \textbf{58.91} & 60.16 & \textbf{29.14} & 49.99 \\
\textbf{\textsc{MeTA-LoRA} (7B)} & \textbf{57.02} & 58.46 & \textbf{79.12} & 23.84 & \textbf{54.61} \\
\midrule
LLaMA* (13B) & 51.04 & 57.88 & 52.97 & 30.41 & 48.08 \\
BX\textsubscript{LLaMA}* (13B) & 53.27 & \textbf{62.12} & 63.65 & \textbf{35.71} & 53.69 \\
\textbf{\textsc{MeTA-LoRA} (13B)} & \textbf{57.87} & 59.58 & \textbf{82.18} & 23.84 & \textbf{55.87} \\
\bottomrule
\end{tabular}
\end{center}
\label{main: multilingual-results}
\end{table}

\begin{table*}[!t]
  \caption{Performance comparison on the more challenging five-task dataset, where PiQA is replaced by WinoGrande. \textsc{MeTA-LoRA} fine-tunes LLaMA2-7B and LLaMA2-13B using 50 examples per task, while LoRA and HydraLoRA correspond to fine-tuning on the entire datatset.}
  \begin{center}
  \renewcommand{\arraystretch}{0.9}
  \small
  \resizebox{\textwidth}{!}{\begin{tabular}{m{70pt}<{\centering}|m{65pt}<{\centering}|m{65pt}<{\centering}|m{65pt}<{\centering}|m{65pt}<{\centering}}
    \toprule
    \textbf{Method} & \textbf{MMLU} & \textbf{MMLU-math} & \textbf{BBH}  & \textbf{AVG} \\
    \toprule
    
    \multicolumn{5}{c}{LLaMA2-7B} \\
    \midrule

    LoRA               & 39.03 & 29.16 & 32.01 & 33.40 \\
    HydraLoRA          & 47.19 & 30.98 & 36.34 & 38.17 \\
    \textsc{MeTA-LoRA} & 46.50 & 32.15 & 39.07 & \textbf{39.24} \\
    \midrule
    
    \multicolumn{5}{c}{LLaMA2-13B} \\
    \midrule

    LoRA               & 39.17 & 28.06 & 43.44 & 36.89 \\
    HydraLoRA          & 54.12 & 30.41 & 44.28 & 42.94 \\
    \textsc{MeTA-LoRA} & 54.18 & 30.51 & 46.02 & \textbf{43.57} \\
    
    \bottomrule
  \end{tabular}}
  \end{center}
  \label{abalation: 5-tasks-harder-results}
\end{table*}

\subsubsection{Performance on Multilingual Learning}
The average performance across all languages for XCOPA, XStoryCloze, XWinograd, and EXAMS is reported in Table~\ref{main: multilingual-results}. Notably, even with only 50 randomly sampled examples per language, \textsc{MeTA-LoRA} achieves competitive or superior performance compared to the BX models, particularly on XCOPA and XWinograd. Moreover, it significantly outperforms the vanilla LLaMA2 models without multilingual adaptation on three out of four benchmarks. This substantial improvement in language understanding can be attributed to the \textsc{MeTA-LoRA} mechanism, which enables rapid adaptation across diverse languages using only a limited amount of data. By leveraging its two-stage optimization framework, \textsc{MeTA-LoRA} effectively captures both shared multilingual knowledge and language-specific patterns, thereby achieving strong data efficiency in multilingual adaptation.

    


    
      


    

\begin{figure}[t]
    \centering
    \begin{subfigure}[t]{0.49\textwidth}
        \centering
        \includegraphics[width=\linewidth]{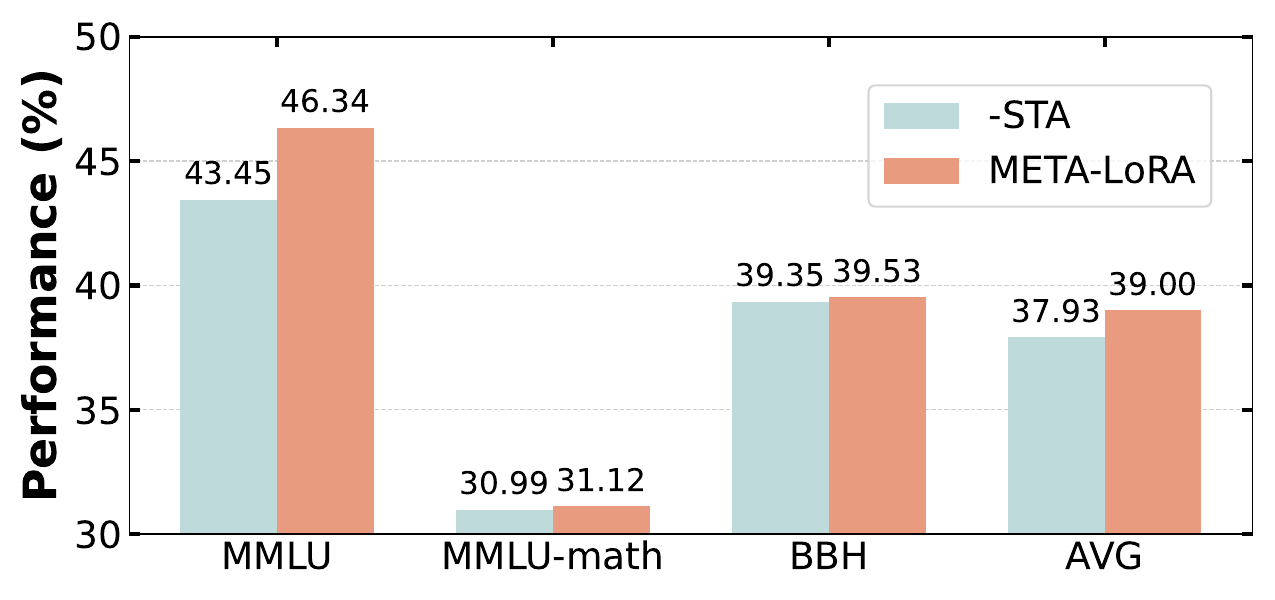}
        \caption{LLaMA2-7B, 50 examples per task.}
        \label{fig:ablation_50_7B}
    \end{subfigure}
    \hfill
    \begin{subfigure}[t]{0.49\textwidth}
        \centering
        \includegraphics[width=\linewidth]{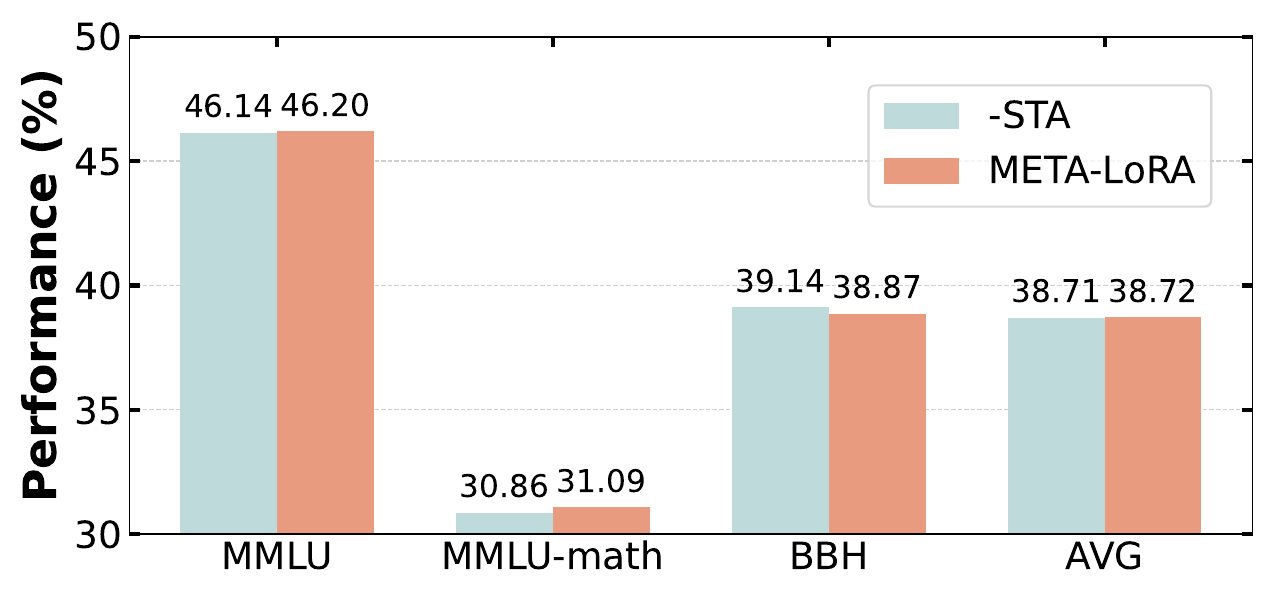}
        \caption{LLaMA2-7B, 100 examples per task.}
        \label{fig:ablation_100_7B}
    \end{subfigure}

    \vskip\baselineskip   

    \begin{subfigure}[t]{0.49\textwidth}
        \centering
        \includegraphics[width=\linewidth]{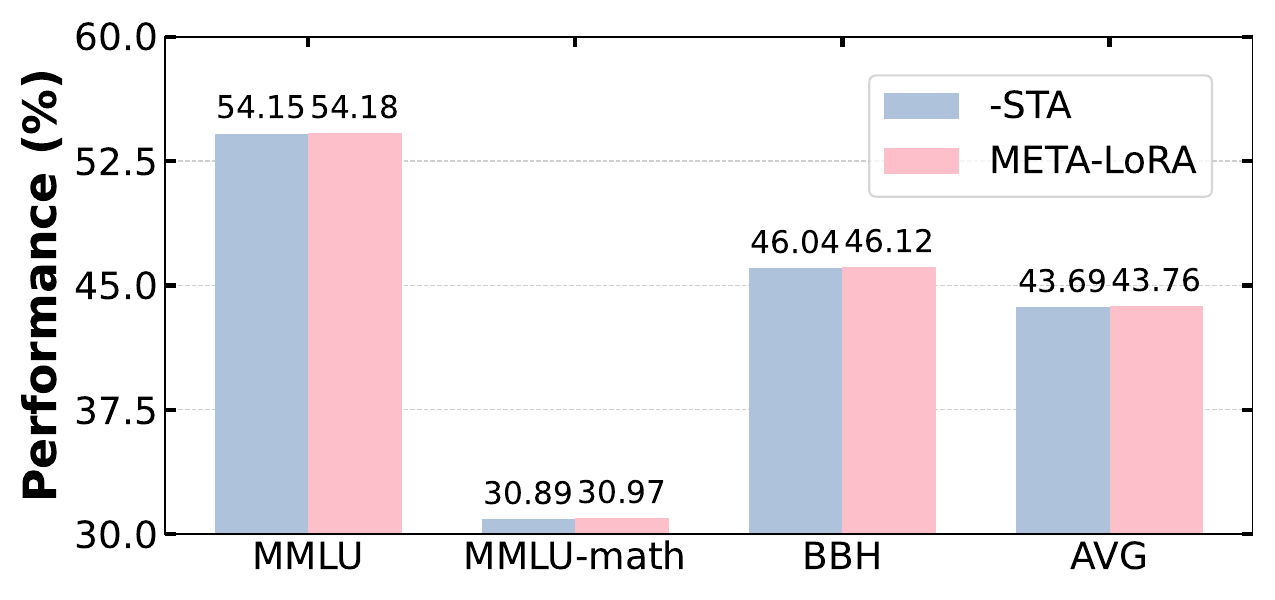}
        \caption{LLaMA2-13B, 50 examples per task.}
        \label{fig:ablation_50_13B}
    \end{subfigure}
    \hfill
    \begin{subfigure}[t]{0.49\textwidth}
        \centering
        \includegraphics[width=\linewidth]{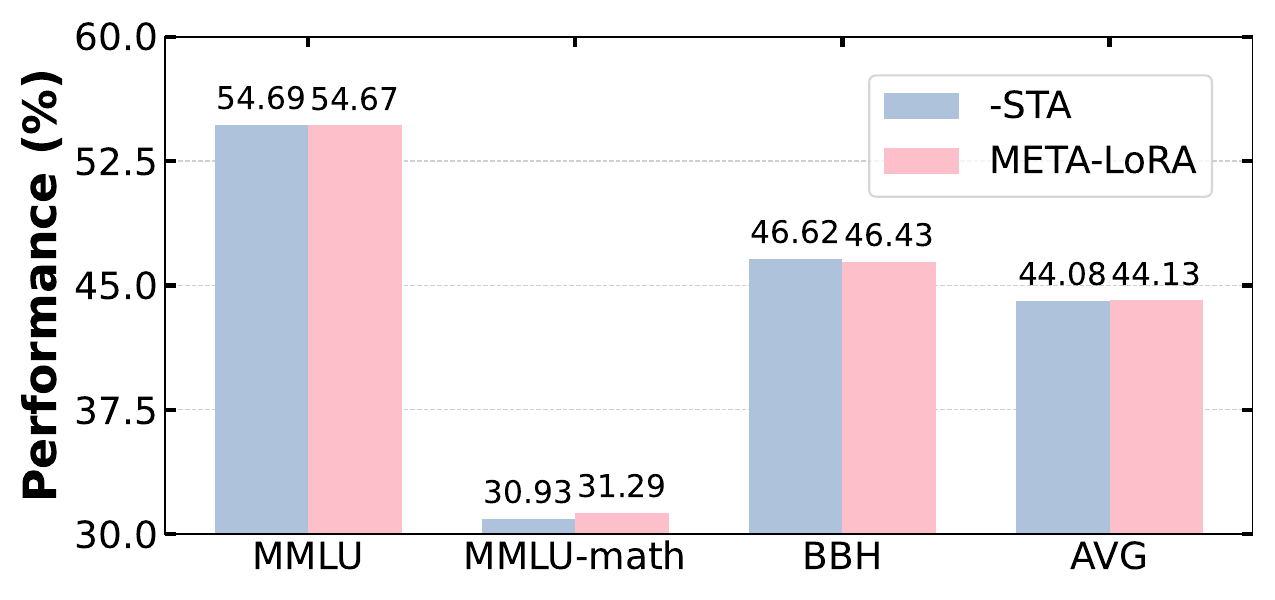}
        \caption{LLaMA2-13B, 100 examples per task.}
        \label{fig:ablation_100_13B}
    \end{subfigure}

    \caption{Ablation results on LLaMA2-7B and LLaMA2-13B under the five-task learning setting. \textbf{-STA} refers to the variant without the task-specific adaptation stage, highlighting its contribution to overall performance.}
    \label{abalation: 5tasks}
\end{figure}

\subsection{Ablation Studies}
\textbf{RQ1: Do more challenging datasets better reveal the generation ability of models?}

To further assess the effectiveness of \textsc{MeTA-LoRA} in more challenging multi-task learning scenarios, we construct a new five-task configuration by replacing PiQA with WinoGrande~\citep{sakaguchi2021winogrande}. Compared to PiQA, WinoGrande is a larger and more difficult commonsense reasoning dataset. It features cloze-style questions in which the model is required to select the correct option from two candidates to complete a given sentence.
We adopt LLaMA2-7B and LLaMA2-13B as base models.  LoRA and HydraLoRA serve as the baselines by fine-tuning the base models on the full dataset, while \textsc{MeTA-LoRA} uses only 50 examples per task. The evaluation results on MMLU, MMLU-math, and BBH are summarized in Table~\ref{abalation: 5-tasks-harder-results}. 

It can be observed that, compared to the standard five-task configuration (as shown in Table~\ref{main: 5-tasks-results}), fine-tuning LLaMA2-7B on the more challenging task sets using LoRA, HydraLoRA or \textsc{MeTA-LoRA} leads to a moderate improvement in the reasoning ability of the model. Specifically, LoRA, HydraLoRA and \textsc{MeTA-LoRA} achieve average performance gains of 4.70\%, 0.14\% and 0.24\%, respectively. For LLaMA2-13B, although overall performance tends to saturate due to the strong inherent generalization of the model, \textsc{MeTA-LoRA} continues to substantially outperform LoRA and HydraLoRA across configurations, highlighting the robustness of \textsc{MeTA-LoRA} under more challenging task distributions and its scalability to larger models.


\textbf{RQ2: Is the two-stage optimization framework necessary for the fine-tuning process?}

To further understand the contribution of the task-specific adaptation stage in \textsc{MeTA-LoRA}, we conduct an ablation study, with the results presented in Figure~\ref{abalation: 5tasks}. Specifically, we investigate a variant, denoted as \textsc{-STA}, in which the task-specific adaptation stage is removed. In this variant, only the gradients computed from the query sets of multiple selected tasks are aggregated to update the shared LoRA adapters in each iteration. 
Experiments are performed using two backbone models, LLaMA2-7B and LLaMA2-13B, under the proposed five-task learning configuration. For each task, either 50 or 100 examples are randomly selected for fine-tuning. In addition, both \textsc{-STA} and \textsc{MeTA-LoRA} are evaluated on three benchmarks: MMLU, the mathematics tasks of MMLU and BBH.


As shown in Figure~\ref{abalation: 5tasks}, the task-specific adaptation stage contributes positively to model performance across both model scales and data regimes, illustrating the necessity and effectiveness of the two-stage design. Notably, the performance gains introduced by the task-specific adaptation stage are more pronounced when using the smaller model (LLaMA2-7B) and the more limited amount of fine-tuning data (50 examples per task). In this case, \textsc{MeTA-LoRA} improves the average score by 1.07\%, demonstrating its ability to capture more task-specific knowledge prior to meta-aggregation. Moreover, while the larger LLaMA2-13B model already exhibits strong generalization capabilities even with a simplified optimization process, the two-stage structure continues to deliver consistent performance benefits, confirming its robustness and scalability across model sizes.

\begin{table*}[!t]
  \caption{Comparison of LoRA and \textsc{MeTA-LoRA} on the standard five-task learning scenario with varying amounts of fine-tuning data per task.}
  \begin{center}
  \renewcommand{\arraystretch}{0.9}
  \small
  \begin{tabular}{m{50pt}<{\centering}|m{60pt}<{\centering}|m{50pt}<{\centering}|m{55pt}<{\centering}|m{50pt}<{\centering}|m{50pt}<{\centering}}
    \toprule
    
    \# Data / task & Method & MMLU & MMLU-math & BBH  & AVG \\
    \toprule

    \multicolumn{6}{c}{LLaMA2-7B} \\
    \midrule

    \multirow{3}{*}{50}
      & LoRA                     & 46.60 & 32.60 & 31.06 & 36.75 \\
      & HydraLoRA                & 46.21 & 29.42 & 37.30 & 37.64 \\
      & \textsc{MeTA-LoRA}       & 46.34 & 31.12 & 39.53 & \textbf{39.00} \\      
    \midrule

    \multirow{3}{*}{100}
      & LoRA                 & 46.49 & 27.32 & 37.17 & 36.99 \\
      & HydraLoRA            & 46.72 & 29.78 & 37.44 & 37.98 \\
      & \textsc{MeTA-LoRA}   & 46.20 & 31.09 & 38.87 & \textbf{38.72} \\
      
    \midrule
    \multicolumn{6}{c}{LLaMA2-13B} \\
    \midrule

    \multirow{3}{*}{50}
      & LoRA                     & 54.17 & 30.47 & 45.89 & 43.51 \\
      & HydraLoRA                & 52.82 & 31.22 & 34.80 & 39.61 \\
      & \textsc{MeTA-LoRA}       & 54.18 & 30.97 & 46.12 & \textbf{43.76} \\      
    \midrule

    \multirow{3}{*}{100}
      & LoRA                 & 55.12 & 31.11 & 45.96 & 44.06 \\
      & HydraLoRA            & 54.11 & 31.30 & 40.50 & 41.97 \\
      & \textsc{MeTA-LoRA}   & 54.67 & 31.29 & 46.43 & \textbf{44.13} \\
    
    \bottomrule
  \end{tabular}
  \end{center}
  \label{PA: 5tasks-num}
\end{table*}

\subsection{Parameter Analysis}

\textbf{RQ3: Does scaling the fine-tuning data always improve performance in multi-task learning?}

As shown in Table~\ref{PA: 5tasks-num}, we analyze how the scale of fine-tuning data impacts model performance under the standard five-task learning setting. Specifically, we fine-tune LLaMA2-7B and LLaMA2-13B using LoRA, HydraLoRA and \textsc{MeTA-LoRA} with either 50 or 100 examples per task. Subsequently, we evaluate the overall performance on MMLU, MMLU-math and BBH. Additional results on the more challenging variant of this setting are reported in Table~\ref{PA: 5tasks-hard-num} of Appendix~\ref{appendix:data_analysis}.

Several insights can be drawn from the results. Overall, increasing the number of fine-tuning examples per task generally improves performance. Moreover, \textsc{MeTA-LoRA} consistently achieves higher average scores across model sizes and data regimes compared to both LoRA and HydraLoRA. For example, under the standard five-task setting with only 50 examples per task, \textsc{MeTA-LoRA} surpasses LoRA and HydraLoRA on LLaMA2-7B by 2.25\% and 1.36\% in average score, respectively. This advantage stems from its novel two-stage optimization framework, which enables strong gains in low-resource settings by extracting task-specific knowledge from only a few examples, while also maintaining stable improvements at larger data scales by mitigating task conflicts. 

Interestingly, LoRA sometimes performs better with only 50 or 100 examples per task than with the full dataset. This counter-intuitive result arises from architectural limitations in multi-task learning. LoRA uses a single shared low-rank adapter that struggles to capture the heterogeneous requirements of diverse tasks. As the data scale grows, task-specific signals interfere with one another, leading to degraded performance. HydraLoRA alleviates this issue to some extent with its asymmetric design.

\section{Conclusion}
In this work, we introduce \textsc{MeTA-LoRA}, a simple yet effective two-stage optimization framework designed to enhance data efficiency in multi-task adaptation of LLMs. By explicitly decoupling task-specific adaptation and meta-knowledge aggregation, \textsc{MeTA-LoRA} is able to quickly adapt to individual tasks using only a few examples, while simultaneously promoting cross-task generalization through shared parameter updates. Comprehensive experiments across both multi-task and multilingual learning settings demonstrate \textsc{MeTA-LoRA} consistently matches or outperforms standard full-data LoRA fine-tuning, despite using significantly less task-specific supervision. These results highlight the potential of \textsc{MeTA-LoRA} as a practical and scalable solution for efficient fine-tuning in real-world low-resource and multi-task scenarios.

\bibliography{references}

\begin{thebibliography}{10}

\bibitem{brown2020language}
Tom Brown, Benjamin Mann, Nick Ryder, Melanie Subbiah, Jared~D Kaplan, Prafulla Dhariwal, Arvind Neelakantan, Pranav Shyam, Girish Sastry, Amanda Askell, et~al.
\newblock Language models are few-shot learners.
\newblock {\em Advances in neural information processing systems}, 33:1877--1901, 2020.

\bibitem{devlin2019bert}
Jacob Devlin, Ming-Wei Chang, Kenton Lee, and Kristina Toutanova.
\newblock Bert: Pre-training of deep bidirectional transformers for language understanding.
\newblock In {\em Proceedings of the 2019 conference of the North American chapter of the association for computational linguistics: human language technologies, volume 1 (long and short papers)}, pages 4171--4186, 2019.

\bibitem{chang2024survey}
Yupeng Chang, Xu~Wang, Jindong Wang, Yuan Wu, Linyi Yang, Kaijie Zhu, Hao Chen, Xiaoyuan Yi, Cunxiang Wang, Yidong Wang, et~al.
\newblock A survey on evaluation of large language models.
\newblock {\em ACM transactions on intelligent systems and technology}, 15(3):1--45, 2024.

\bibitem{han2024parameter}
Zeyu Han, Chao Gao, Jinyang Liu, Jeff Zhang, and Sai~Qian Zhang.
\newblock Parameter-efficient fine-tuning for large models: A comprehensive survey.
\newblock {\em arXiv preprint arXiv:2403.14608}, 2024.

\bibitem{hu2022lora}
Edward~J Hu, Yelong Shen, Phillip Wallis, Zeyuan Allen-Zhu, Yuanzhi Li, Shean Wang, Lu~Wang, Weizhu Chen, et~al.
\newblock Lora: Low-rank adaptation of large language models.
\newblock {\em ICLR}, 1(2):3, 2022.

\bibitem{rebuffi2017adapters}
Sylvestre-Alvise Rebuffi, Hakan Bilen, and Andrea Vedaldi.
\newblock Learning multiple visual domains with residual adapters.
\newblock {\em Advances in neural information processing systems}, 30, 2017.

\bibitem{li2021prefix}
Xiang~Lisa Li and Percy Liang.
\newblock Prefix-tuning: Optimizing continuous prompts for generation.
\newblock {\em arXiv preprint arXiv:2101.00190}, 2021.

\bibitem{chang2024ba}
Yupeng Chang, Yi~Chang, and Yuan Wu.
\newblock Ba-lora: Bias-alleviating low-rank adaptation to mitigate catastrophic inheritance in large language models.
\newblock {\em arXiv preprint arXiv:2408.04556}, 2024.

\bibitem{liu2025rlora}
Jinda Liu, Yi~Chang, and Yuan Wu.
\newblock R-lora: Random initialization of multi-head lora for multi-task learning.
\newblock {\em arXiv preprint arXiv:2502.15455}, 2025.

\bibitem{tian2024hydralora}
Chunlin Tian, Zhan Shi, Zhijiang Guo, Li~Li, and Cheng-Zhong Xu.
\newblock Hydralora: An asymmetric lora architecture for efficient fine-tuning.
\newblock {\em Advances in Neural Information Processing Systems}, 37:9565--9584, 2024.

\bibitem{liu2023coreset1}
Wei Liu, Weihao Zeng, Keqing He, Yong Jiang, and Junxian He.
\newblock What makes good data for alignment? a comprehensive study of automatic data selection in instruction tuning.
\newblock {\em arXiv preprint arXiv:2312.15685}, 2023.

\bibitem{xia2024less}
Mengzhou Xia, Sadhika Malladi, Suchin Gururangan, Sanjeev Arora, and Danqi Chen.
\newblock Less: Selecting influential data for targeted instruction tuning.
\newblock {\em arXiv preprint arXiv:2402.04333}, 2024.

\bibitem{azeemi2023dataprunig}
Abdul~Hameed Azeemi, Ihsan Qazi, and Agha~Ali Raza.
\newblock Data pruning for efficient model pruning in neural machine translation.
\newblock In {\em Findings of the Association for Computational Linguistics: EMNLP 2023}, pages 236--246, 2023.

\bibitem{lin2024re}
Xinyu Lin, Wenjie Wang, Yongqi Li, Shuo Yang, Fuli Feng, Yinwei Wei, and Tat-Seng Chua.
\newblock Data-efficient fine-tuning for llm-based recommendation.
\newblock In {\em Proceedings of the 47th international ACM SIGIR conference on research and development in information retrieval}, pages 365--374, 2024.

\bibitem{chen2024molecular}
Dingshuo Chen, Zhixun Li, Yuyan Ni, Guibin Zhang, Ding Wang, Qiang Liu, Shu Wu, Jeffrey Yu, and Liang Wang.
\newblock Beyond efficiency: Molecular data pruning for enhanced generalization.
\newblock {\em Advances in Neural Information Processing Systems}, 37:18036--18061, 2024.

\bibitem{finn2017maml}
Chelsea Finn, Pieter Abbeel, and Sergey Levine.
\newblock Model-agnostic meta-learning for fast adaptation of deep networks.
\newblock In {\em International conference on machine learning}, pages 1126--1135. PMLR, 2017.

\bibitem{rebuffi2017learning}
Sylvestre-Alvise Rebuffi, Hakan Bilen, and Andrea Vedaldi.
\newblock Learning multiple visual domains with residual adapters.
\newblock {\em Advances in neural information processing systems}, 30, 2017.

\bibitem{houlsby2019parameter}
Neil Houlsby, Andrei Giurgiu, Stanislaw Jastrzebski, Bruna Morrone, Quentin De~Laroussilhe, Andrea Gesmundo, Mona Attariyan, and Sylvain Gelly.
\newblock Parameter-efficient transfer learning for nlp.
\newblock In {\em International conference on machine learning}, pages 2790--2799. PMLR, 2019.

\bibitem{sung2022vl}
Yi-Lin Sung, Jaemin Cho, and Mohit Bansal.
\newblock Vl-adapter: Parameter-efficient transfer learning for vision-and-language tasks.
\newblock In {\em Proceedings of the IEEE/CVF conference on computer vision and pattern recognition}, pages 5227--5237, 2022.

\bibitem{stickland2019bert}
Asa~Cooper Stickland and Iain Murray.
\newblock Bert and pals: Projected attention layers for efficient adaptation in multi-task learning.
\newblock In {\em International Conference on Machine Learning}, pages 5986--5995. PMLR, 2019.

\bibitem{liu2024gpt}
Xiao Liu, Yanan Zheng, Zhengxiao Du, Ming Ding, Yujie Qian, Zhilin Yang, and Jie Tang.
\newblock Gpt understands, too.
\newblock {\em AI Open}, 5:208--215, 2024.

\bibitem{lester2021power}
Brian Lester, Rami Al-Rfou, and Noah Constant.
\newblock The power of scale for parameter-efficient prompt tuning.
\newblock {\em arXiv preprint arXiv:2104.08691}, 2021.

\bibitem{liu2021p}
Xiao Liu, Kaixuan Ji, Yicheng Fu, Weng~Lam Tam, Zhengxiao Du, Zhilin Yang, and Jie Tang.
\newblock P-tuning v2: Prompt tuning can be comparable to fine-tuning universally across scales and tasks.
\newblock {\em arXiv preprint arXiv:2110.07602}, 2021.

\bibitem{sung2021training}
Yi-Lin Sung, Varun Nair, and Colin~A Raffel.
\newblock Training neural networks with fixed sparse masks.
\newblock {\em Advances in Neural Information Processing Systems}, 34:24193--24205, 2021.

\bibitem{dey2024sparse}
Nolan Dey, Shane Bergsma, and Joel Hestness.
\newblock Sparse maximal update parameterization: A holistic approach to sparse training dynamics.
\newblock {\em arXiv preprint arXiv:2405.15743}, 2024.

\bibitem{touvron2023llama}
Hugo Touvron, Thibaut Lavril, Gautier Izacard, Xavier Martinet, Marie-Anne Lachaux, Timoth{\'e}e Lacroix, Baptiste Rozi{\`e}re, Naman Goyal, Eric Hambro, Faisal Azhar, et~al.
\newblock Llama: Open and efficient foundation language models.
\newblock {\em arXiv preprint arXiv:2302.13971}, 2023.

\bibitem{huang2023lorahub}
Chengsong Huang, Qian Liu, Bill~Yuchen Lin, Tianyu Pang, Chao Du, and Min Lin.
\newblock Lorahub: Efficient cross-task generalization via dynamic lora composition.
\newblock {\em arXiv preprint arXiv:2307.13269}, 2023.

\bibitem{wang2023multilora}
Yiming Wang, Yu~Lin, Xiaodong Zeng, and Guannan Zhang.
\newblock Multilora: Democratizing lora for better multi-task learning.
\newblock {\em arXiv preprint arXiv:2311.11501}, 2023.

\bibitem{zadouri2023pushing}
Ted Zadouri, Ahmet {\"U}st{\"u}n, Arash Ahmadian, Beyza Ermi{\c{s}}, Acyr Locatelli, and Sara Hooker.
\newblock Pushing mixture of experts to the limit: Extremely parameter efficient moe for instruction tuning.
\newblock {\em arXiv preprint arXiv:2309.05444}, 2023.

\bibitem{dou2023loramoe}
Shihan Dou, Enyu Zhou, Yan Liu, Songyang Gao, Jun Zhao, Wei Shen, Yuhao Zhou, Zhiheng Xi, Xiao Wang, Xiaoran Fan, et~al.
\newblock Loramoe: Revolutionizing mixture of experts for maintaining world knowledge in language model alignment.
\newblock {\em arXiv preprint arXiv:2312.09979}, 4(7), 2023.

\bibitem{ding2023efficiency}
Tianyu Ding, Tianyi Chen, Haidong Zhu, Jiachen Jiang, Yiqi Zhong, Jinxin Zhou, Guangzhi Wang, Zhihui Zhu, Ilya Zharkov, and Luming Liang.
\newblock The efficiency spectrum of large language models: An algorithmic survey.
\newblock {\em arXiv preprint arXiv:2312.00678}, 2023.

\bibitem{xu2024survey}
Mengwei Xu, Wangsong Yin, Dongqi Cai, Rongjie Yi, Daliang Xu, Qipeng Wang, Bingyang Wu, Yihao Zhao, Chen Yang, Shihe Wang, et~al.
\newblock A survey of resource-efficient llm and multimodal foundation models.
\newblock {\em arXiv preprint arXiv:2401.08092}, 2024.

\bibitem{lin2024data}
Xinyu Lin, Wenjie Wang, Yongqi Li, Shuo Yang, Fuli Feng, Yinwei Wei, and Tat-Seng Chua.
\newblock Data-efficient fine-tuning for llm-based recommendation.
\newblock In {\em Proceedings of the 47th international ACM SIGIR conference on research and development in information retrieval}, pages 365--374, 2024.

\bibitem{xia2022moderate}
Xiaobo Xia, Jiale Liu, Jun Yu, Xu~Shen, Bo~Han, and Tongliang Liu.
\newblock Moderate coreset: A universal method of data selection for real-world data-efficient deep learning.
\newblock In {\em The Eleventh International Conference on Learning Representations}, 2022.

\bibitem{marion2023less}
Max Marion, Ahmet {\"U}st{\"u}n, Luiza Pozzobon, Alex Wang, Marzieh Fadaee, and Sara Hooker.
\newblock When less is more: Investigating data pruning for pretraining llms at scale.
\newblock {\em arXiv preprint arXiv:2309.04564}, 2023.

\bibitem{sorscher2022beyond}
Ben Sorscher, Robert Geirhos, Shashank Shekhar, Surya Ganguli, and Ari Morcos.
\newblock Beyond neural scaling laws: beating power law scaling via data pruning.
\newblock {\em Advances in Neural Information Processing Systems}, 35:19523--19536, 2022.

\bibitem{wei2021flan-v2}
Jason Wei, Maarten Bosma, Vincent~Y Zhao, Kelvin Guu, Adams~Wei Yu, Brian Lester, Nan Du, Andrew~M Dai, and Quoc~V Le.
\newblock Finetuned language models are zero-shot learners.
\newblock {\em arXiv preprint arXiv:2109.01652}, 2021.

\bibitem{suzgun2022bbh}
Mirac Suzgun, Nathan Scales, Nathanael Sch{\"a}rli, Sebastian Gehrmann, Yi~Tay, Hyung~Won Chung, Aakanksha Chowdhery, Quoc~V Le, Ed~H Chi, Denny Zhou, et~al.
\newblock Challenging big-bench tasks and whether chain-of-thought can solve them.
\newblock {\em arXiv preprint arXiv:2210.09261}, 2022.

\bibitem{liu2024moe}
Qidong Liu, Xian Wu, Xiangyu Zhao, Yuanshao Zhu, Derong Xu, Feng Tian, and Yefeng Zheng.
\newblock When moe meets llms: Parameter efficient fine-tuning for multi-task medical applications.
\newblock In {\em Proceedings of the 47th International ACM SIGIR Conference on Research and Development in Information Retrieval}, pages 1104--1114, 2024.

\bibitem{cobbe2021gsm8k}
Karl Cobbe, Vineet Kosaraju, Mohammad Bavarian, Mark Chen, Heewoo Jun, Lukasz Kaiser, Matthias Plappert, Jerry Tworek, Jacob Hilton, Reiichiro Nakano, et~al.
\newblock Training verifiers to solve math word problems.
\newblock {\em arXiv preprint arXiv:2110.14168}, 2021.

\bibitem{wang2017qqp}
Zhiguo Wang, Wael Hamza, and Radu Florian.
\newblock Bilateral multi-perspective matching for natural language sentences.
\newblock {\em arXiv preprint arXiv:1702.03814}, 2017.

\bibitem{huang2019cosmosqa}
Lifu Huang, Ronan~Le Bras, Chandra Bhagavatula, and Yejin Choi.
\newblock Cosmos qa: Machine reading comprehension with contextual commonsense reasoning.
\newblock {\em arXiv preprint arXiv:1909.00277}, 2019.

\bibitem{sap2019socialiqa}
Maarten Sap, Hannah Rashkin, Derek Chen, Ronan LeBras, and Yejin Choi.
\newblock Socialiqa: Commonsense reasoning about social interactions.
\newblock {\em arXiv preprint arXiv:1904.09728}, 2019.

\bibitem{bisk2020piqa}
Yonatan Bisk, Rowan Zellers, Jianfeng Gao, Yejin Choi, et~al.
\newblock Piqa: Reasoning about physical commonsense in natural language.
\newblock In {\em Proceedings of the AAAI conference on artificial intelligence}, volume~34, pages 7432--7439, 2020.

\bibitem{hendrycks2020mmlu}
Dan Hendrycks, Collin Burns, Steven Basart, Andy Zou, Mantas Mazeika, Dawn Song, and Jacob Steinhardt.
\newblock Measuring massive multitask language understanding.
\newblock {\em arXiv preprint arXiv:2009.03300}, 2020.

\bibitem{2025-malora}
Xujia Wang, Haiyan Zhao, Shuo Wang, Hanqing Wang, and Zhiyuan Liu.
\newblock {MAL}o{RA}: Mixture of asymmetric low-rank adaptation for enhanced multi-task learning.
\newblock In Luis Chiruzzo, Alan Ritter, and Lu~Wang, editors, {\em Findings of the Association for Computational Linguistics: NAACL 2025}, pages 5609--5626, Albuquerque, New Mexico, April 2025. Association for Computational Linguistics.

\bibitem{li2023bactrian}
Haonan Li, Fajri Koto, Minghao Wu, Alham~Fikri Aji, and Timothy Baldwin.
\newblock Bactrian-x: Multilingual replicable instruction-following models with low-rank adaptation.
\newblock {\em arXiv preprint arXiv:2305.15011}, 2023.

\bibitem{taori2023alpaca}
Rohan Taori, Ishaan Gulrajani, Tianyi Zhang, Yann Dubois, Xuechen Li, Carlos Guestrin, Percy Liang, and Tatsunori~B Hashimoto.
\newblock Stanford alpaca: An instruction-following llama model, 2023.

\bibitem{conover2023dolly}
Mike Conover, Matt Hayes, Ankit Mathur, Jianwei Xie, Jun Wan, Sam Shah, Ali Ghodsi, Patrick Wendell, Matei Zaharia, and Reynold Xin.
\newblock Free dolly: Introducing the world’s first truly open instruction-tuned llm, 2023.

\bibitem{ponti2020xcopa}
Edoardo~Maria Ponti, Goran Glava{\v{s}}, Olga Majewska, Qianchu Liu, Ivan Vuli{\'c}, and Anna Korhonen.
\newblock Xcopa: A multilingual dataset for causal commonsense reasoning.
\newblock {\em arXiv preprint arXiv:2005.00333}, 2020.

\bibitem{lin2022XStoryCloze}
Xi~Victoria Lin, Todor Mihaylov, Mikel Artetxe, Tianlu Wang, Shuohui Chen, Daniel Simig, Myle Ott, Naman Goyal, Shruti Bhosale, Jingfei Du, et~al.
\newblock Few-shot learning with multilingual generative language models.
\newblock In {\em Proceedings of the 2022 conference on empirical methods in natural language processing}, pages 9019--9052, 2022.

\bibitem{mostafazadeh2016corpus}
Nasrin Mostafazadeh, Nathanael Chambers, Xiaodong He, Devi Parikh, Dhruv Batra, Lucy Vanderwende, Pushmeet Kohli, and James Allen.
\newblock A corpus and cloze evaluation for deeper understanding of commonsense stories.
\newblock In {\em Proceedings of the 2016 Conference of the North American Chapter of the Association for Computational Linguistics: Human Language Technologies}, pages 839--849, 2016.

\bibitem{XWinoGrad}
Alexey Tikhonov and Max Ryabinin.
\newblock It's all in the heads: Using attention heads as a baseline for cross-lingual transfer in commonsense reasoning.
\newblock {\em arXiv preprint arXiv:2106.12066}, 2021.

\bibitem{XWinoGrad_crosslingual}
Niklas Muennighoff, Thomas Wang, Lintang Sutawika, Adam Roberts, Stella Biderman, Teven~Le Scao, M~Saiful Bari, Sheng Shen, Zheng-Xin Yong, Hailey Schoelkopf, et~al.
\newblock Crosslingual generalization through multitask finetuning.
\newblock {\em arXiv preprint arXiv:2211.01786}, 2022.

\bibitem{hardalov2020exams}
Momchil Hardalov, Todor Mihaylov, Dimitrina Zlatkova, Yoan Dinkov, Ivan Koychev, and Preslav Nakov.
\newblock Exams: A multi-subject high school examinations dataset for cross-lingual and multilingual question answering.
\newblock {\em arXiv preprint arXiv:2011.03080}, 2020.

\bibitem{sakaguchi2021winogrande}
Keisuke Sakaguchi, Ronan~Le Bras, Chandra Bhagavatula, and Yejin Choi.
\newblock Winogrande: An adversarial winograd schema challenge at scale.
\newblock {\em Communications of the ACM}, 64(9):99--106, 2021.

\end{thebibliography}
\bibliographystyle{unsrt} 

\appendix
\section{Appendix}
\subsection{Datasets and Hyper-parameters}
\subsubsection{Details for Flanv2 Setting}
\label{appendix:flanv2}
Following~\citep{tian2024hydralora}, we select a portion of the Flanv2 datasets covering Natural Language Understanding (NLU) and Natural Language Generation (NLG), which can be grouped into 10 distinct task clusters. Then we evaluate it with the Big-Bench Hard (BBH) benchmark.

We summarize the details of the used datasets as follows:

\begin{enumerate}
    \item \textbf{Struct-to-Text Conversion}: This task evaluates the capability to generate natural language descriptions from structured data inputs. We use the following datasets: (1) CommonGen; (2) DART; (3) E2ENLG; (4) WebNLG
    
    \item \textbf{Translation}: Translation involves converting text from one language to another, maintaining the original meaning and nuances. We use the following datasets: (1) En-Fr from WMT'14; (2) En-De, En-Tr, En-Ru, En-Fi, En-Ro from WMT'16; (3) En-Es from Paracrawl.
    
    \item \textbf{Commonsense Reasoning}: This involves assessing the ability to apply physical or scientific principles alongside common sense in reasoning tasks. We use the following datasets: (1) COPA; (2) HellaSwag; (3) PiQA; (4) StoryCloze.
    
    \item \textbf{Sentiment Analysis}: A fundamental task in natural language processing (NLP) that determines the sentiment polarity (positive or negative) of a given text. We use the following datasets: (1) IMDB; (2) Sentiment140; (3) SST-2; (4) Yelp.
    
    \item \textbf{Paraphrase Detection}: This task requires models to ascertain whether two sentences convey the same meaning, indicating semantic equivalence. We use the following datasets: (1) MRPC; (2) QQP; (3) Paws Wiki.
    
    \item \textbf{Coreference Resolution}: Involves identifying instances within a text that refer to the same entity, demonstrating an understanding of textual context. We use the following datasets: (1) DPR; (2) WSC273.
    
    \item \textbf{Reading Comprehension}: Assesses the capability to derive answers to questions from a provided text containing relevant information. We use the following datasets: (1) BoolQ; (2) DROP; (3) MultiRC; (4) OBQA; (5) SQuADv1; (6) SQuADv2.
    
    \item \textbf{Reading Comprehension with Commonsense}: Merges traditional reading comprehension skills with commonsense reasoning, requiring understanding beyond the explicit text. We use the following datasets: (1) CosmosQA; (2) ReCoRD.

    \item \textbf{Natural Language Inference}: Focuses on deducing the relationship between two sentences, determining if the second sentence logically follows from, contradicts, or is unrelated to the first sentence. We use the following datasets: (1) ANLI; (2) CB; (3) MNLI; (4) QNLI; (5) SNLI; (6) WNLI; (7) RTE.
    
    \item \textbf{Closed-Book Question Answering}: This task challenges models to answer questions about general knowledge without direct access to external information sources. We use the following datasets: (1) ARC; (2) NQ; (3) TriviaQA.

\end{enumerate}

\subsubsection{Hyper-parameter Settings}
\label{appendix:paras}
Details on hyperparameters used for \textsc{MeTA-LoRA}, LoRA and HydraLoRA are provided below.

\textbf{\textsc{MeTA-LoRA}}: For all experiments, we integrate adapter modules into every dense layer of the multi-head attention (namely $Q$, $K$, $V$, and $O$) in the selected LLMs. Also, we set the low-rank parameter $r$ to $16$.

\textbf{LoRA}:  In all experiments, adapter modules are inserted into every dense layer of the multi-head attention components (namely $Q$, $K$, $V$, and $O$) in the selected LLMs, with the rank $r$ set to $16$. In addition, the learning rate is set to $2 \times 10^{-4}$ and the batch size is set to $8$. For experiments on the full datasets, results are reported after a single epoch of fine-tuning. For parameter analysis with respect to the amount of fine-tuning data, LoRA fine-tunes the base models for 5 epochs when using 50 examples per task, and for 10 epochs when using 100 examples per task. Finally, we report the best results obtained on the benchmarks.

\textbf{HydraLoRA}: For experiments not covered in the original paper~\citep{tian2024hydralora}, we adopt the default configurations suggested by HydraLoRA and adjust the number of $B$ matrices (\texttt{lora\_nums}) to match the number of tasks. Consistent with LoRA, full-dataset fine-tuning is performed for a single epoch. Under limited-data settings, the base models are fine-tuned for 5 epochs with 50 examples per task and for 10 epochs with 100 examples per task. Finally, we report the best results obtained in each setting.

\subsection{Training Efficiency}
To further assess the efficiency of the proposed method, we measure the running time of different fine-tuning schemes under identical experimental environments. As presented in Table~\ref{appendix:time}, \textsc{MeTA-LoRA} substantially reduces training time compared to standard LoRA and advanced HydraLoRA, while maintaining competitive performance.

\textbf{\begin{table}[!t]
\caption{Comparison of running time across fine-tuning schemes. LoRA and HydraLoRA fine-tune the base models on the more challenging dataset for one epoch, while \textsc{MeTA-LoRA} performs 1,000 iterations using 100 samples from each task.}
\begin{center}
\begin{tabular}{m{55pt}<{\centering}|m{30pt}<{\centering}<{\centering}m{55pt}<{\centering}m{70pt}<{\centering}}
\toprule
\textbf{Model Size} & \textbf{LoRA} &\textbf{HydraLoRA} & \textbf{\textsc{MeTA-LoRA}} \\
\midrule
\textsc{7B}   & 17h & 37h & \textbf{3h} \\
\textsc{13B}  & 29h & 57h & \textbf{5h}  \\
\bottomrule
\end{tabular}
\end{center}
\label{appendix:time}
\end{table}
}

\subsection{Parameter Analysis}

\begin{table}[!t]
\caption{Performance comparison on the standard five-task setting with respect to $\alpha$ and $\beta$.}
\begin{center}
\begin{tabular}{lccc}
\toprule
Combination & $\alpha$ & $\beta$ & BBH \\
\midrule
$c_{base}$(ours) & $5 \times 10^{-6}$    & $2 \times 10^{-6}$ & \textbf{39.53} \\
$c_1$ & $2 \times 10^{-6}$    & $5 \times 10^{-6}$ & 39.39 \\
$c_2$ & $5 \times 10^{-5}$    & $2 \times 10^{-5}$ & 37.90 \\
$c_3$ & $5 \times 10^{-7}$    & $2 \times 10^{-7}$ & 38.96 \\
\bottomrule
\end{tabular}
\end{center}
\label{tab:lr_analysis}
\end{table}

\subsubsection{Impact of learning rates}
To investigate the impact of learning rates, we fine-tuned LLaMA2-7B under the standard five-task learning setting using various $(\alpha, \beta)$ configurations. As presented in Table~\ref{tab:lr_analysis}, $c_2$ with higher learning rates and $c_3$ with lower learning rates both degrade overall performance, suggesting optimization instability and insufficient adaptation. 


\begin{table*}[!t]
  \caption{Comparative analysis of LoRA, HydraLoRA and META-LORA on the more challenging five-task learning scenario with varying amounts of fine-tuning data per task.}
  \begin{center}
  \renewcommand{\arraystretch}{0.9}
  \small
  \begin{tabular}{m{50pt}<{\centering}|m{60pt}<{\centering}|m{50pt}<{\centering}|m{55pt}<{\centering}|m{50pt}<{\centering}|m{50pt}<{\centering}}
    \toprule
    
    \# Data / task & Method & MMLU & MMLU-math & BBH  & AVG \\
    \toprule

    \multicolumn{6}{c}{LLaMA2-7B} \\
    \midrule

    \multirow{3}{*}{50}
      & LoRA                     & 47.04 & 31.32 & 34.90 & 37.75 \\
      & HydraLoRA                & 46.30 & 30.17 & 36.64 & 37.70 \\
      & \textsc{MeTA-LoRA}       & 46.50 & 32.15 & 39.07 & \textbf{39.24} \\
    \midrule

    \multirow{3}{*}{100}
      & LoRA                 & 45.68 & 25.94 & 33.75 & 35.12 \\
      & HydraLoRA            & 46.44 & 30.51 & 36.75 & 37.90 \\
      & \textsc{MeTA-LoRA}   & 46.55 & 31.59 & 39.45 & \textbf{39.20} \\
      
    \midrule
    \multicolumn{6}{c}{LLaMA2-13B} \\
    \midrule

    \multirow{3}{*}{50}
      & LoRA                     & 54.14 & 31.69 & 46.18 & \textbf{44.00} \\
      & HydraLoRA                & 53.18 & 31.83 & 36.25 & 40.42 \\
      & \textsc{MeTA-LoRA}       & 54.18 & 30.51 & 46.02 & 43.57 \\      
    \midrule

    \multirow{3}{*}{100}
      & LoRA                 & 54.36 & 29.22 & 46.17 & 43.25 \\
      & HydraLoRA            & 53.43 & 30.90 & 38.35 & 40.89 \\
      & \textsc{MeTA-LoRA}   & 54.23 & 31.21 & 46.78 & \textbf{44.07} \\
    
    \bottomrule
  \end{tabular}
  \end{center}
  \label{PA: 5tasks-hard-num}
\end{table*}

\subsubsection{Impact of the fine-tuning data}
\label{appendix:data_analysis}
As shown in Table~\ref{PA: 5tasks-hard-num}, we also analyze how the scale of fine-tuning data impacts model performance under the more challenging variant of the standard five-task setting. Specifically, we fine-tune LLaMA2-7B and LLaMA2-13B using LoRA, HydraLoRA and META-LORA with either 50 or 100 examples per task. Subsequently, we evaluate the overall performance on MMLU, MMLU-math, and BBH. 

Consistent with the findings under the standard five-task setting, LoRA achieves better performance with limited data than with full-data, reflecting its inherent architectural limitations. When the difficulty of tasks increases, HydraLoRA struggles to capture task-specific knowledge in such low-resource regimes. In contrast, our method built upon a two-stage optimization framework can rapidly adapt in diverse multi-task scenarios, which further demonstrates its robustness and establishes its superiority over existing approaches.


\end{document}